\RequirePackage{fix-cm}
\documentclass[smallextended]{svjour3}       
\smartqed  
\usepackage{graphicx}
\usepackage{epstopdf}
\usepackage{lineno,hyperref}
\usepackage{multirow}
\usepackage{amsmath}
\usepackage{float}
\usepackage{tabularx}
\usepackage{subfig}
\usepackage{caption}
\usepackage{algorithm}
\usepackage{algorithmic}
\usepackage{wasysym}
\usepackage{picins}
\begin{document}
\newcolumntype{C}[1]{>{\centering\arraybackslash}p{#1}}

\title{Eliminating cross-camera bias for vehicle re-identification
}


\author{Jinjia Peng \textsuperscript{1} \and
        Guangqi Jiang \textsuperscript{1}\and
        Dongyan Chen \textsuperscript{1}\and
        Tongtong Zhao\textsuperscript{1} \and
        Huibing Wang \textsuperscript{1,}\textsuperscript{*} \and
        Xianping Fu \textsuperscript{1,}\textsuperscript{2,}\textsuperscript{*}
}


\institute{Jinjia Peng \at
              {jinjiapeng@dlmu.edu.cn}            
           \and
           Guangqi Jiang \at
              {guangqi-j@dlmu.edu.cn}
           \and
           Dongyan Chen \at
              {chendongyan@dlmu.edu.cn}
           \and
           Tongtong Zhao \at
              {zhaotongtong94@163.com}
           \and
           \XBox Huibing Wang \at
              {huibing.wang@dlmu.edu.cn}
           \and
           \XBox Xianping Fu \at
              {fxp@dlmu.edu.cn} \\
           \and
            *\emph{Both are corresponding authors}
            \and
            \textsuperscript{1}\emph{College of Information and Science Technology, Dalian Maritime University, Dalian, Liaoning, 116021, China. }
            \and
            \textsuperscript{2}\emph{Pengcheng Laboratory, Shenzhen, Guangdong, 518055, China}   
}

\date{Received: date / Accepted: date}

\maketitle

\begin{abstract}
Vehicle re-identification (reID) often requires recognize a target vehicle in large datasets captured from multi-cameras. It plays an important role in the automatic analysis of the increasing urban surveillance videos, which has become a hot topic in recent years.  However, the appearance of vehicle images is easily affected by the environment that various illuminations, different backgrounds and viewpoints, which leads to the large bias between different cameras. To address this problem, this paper proposes a cross-camera adaptation framework (CCA), which smooths the bias by exploiting the common space between cameras for all samples. CCA first transfers images from multi-cameras into one camera to reduce the impact of the illumination and resolution, which generates the samples with the similar distribution. Then, to eliminate the influence of background and focus on the valuable parts, we propose an attention alignment network (AANet) to learn powerful features for vehicle reID. Specially, in AANet, the spatial transfer network with attention module is introduced to locate a series of the most discriminative regions with high-attention weights and suppress the background. Moreover, comprehensive experimental results have demonstrated that our proposed CCA can achieve excellent performances on benchmark datasets VehicleID and VeRi-776.


\end{abstract}

\section{Introduction}

The research related to vehicles has attracted wide attention and made some progress in the field of computer vision, such as vehicle detection \cite{hu2018sinet,ref_article2}, tracking \cite{tang2019cityflow,fang2019road} and classification \cite{ref_article5,ma2019fine}. Different from the tasks above, the purpose of vehicle reID is to accurately match the target vehicle captured from multiple non-overlapping cameras, which is of great significance to intelligent transportation. Meanwhile, the large amount of video or images could be processed automatically carried out by vehicle reID to exploit the meaningful information, which plays an important role in modern smart surveillance systems.

\begin{figure}
\centering
  \includegraphics[width=10cm]{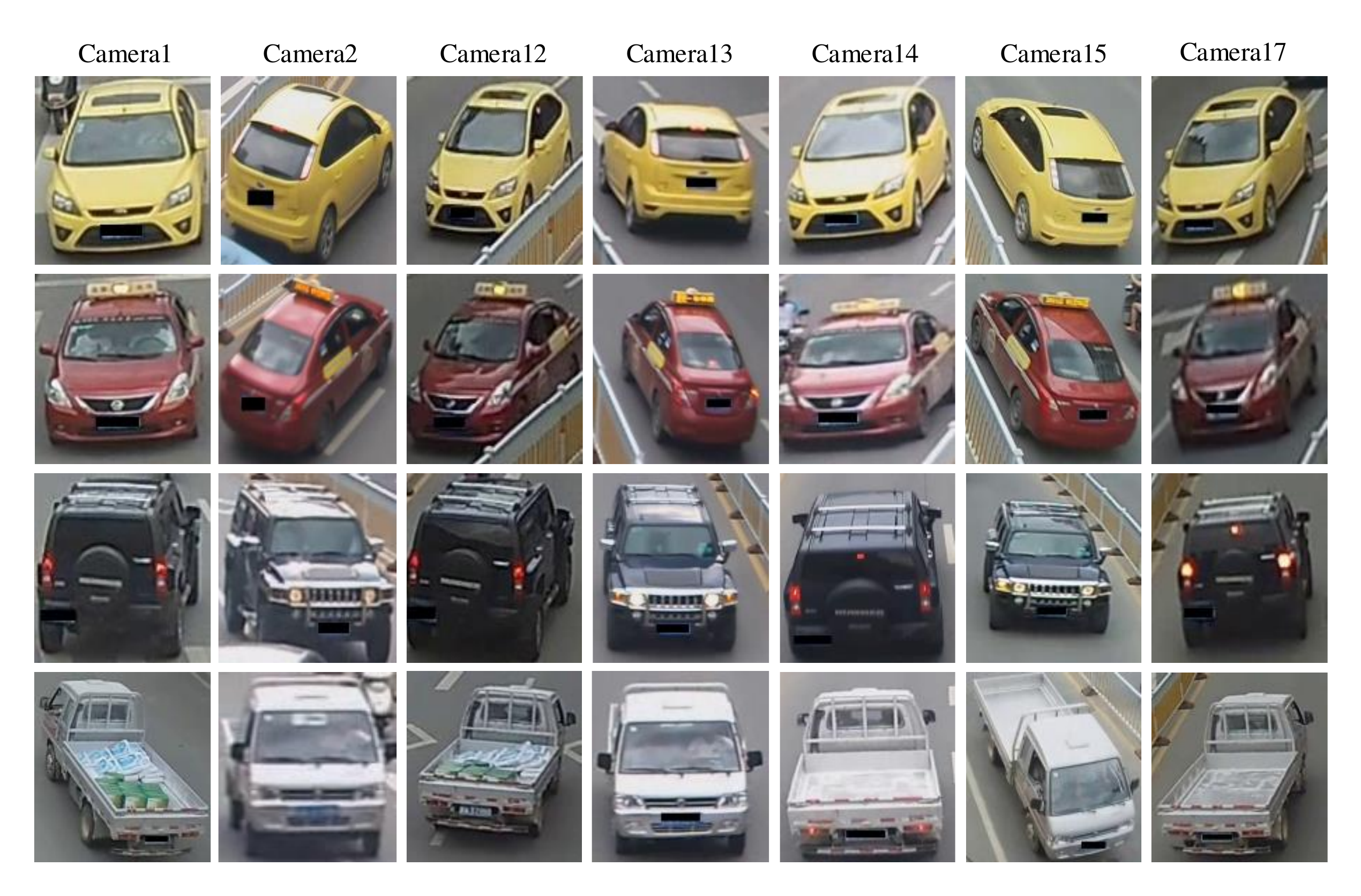}
\caption{Example images from 1,2,12,13,14,15,17 cameras of VeRi-776. The images with a same identity have different appearances in different camera views.}
\label{introduction}       
\end{figure}

With the recent development of deep learning, lots of excellent deep learning-based methods \cite{khorramshahi2019dual,he2019part,zhu2019vehicle,guo2018learning} are proposed for the vehicle reID task. However, there still exist many limitations for the application in the real-world. Different with the person reID \cite{wu2018and,wu20193,wu2019cross,li2017learning,Wu2018where} and fine-grained classification \cite{wang2018multiview,yang2018learning,wu2018deep,wang2017effective,wu2018cycle} that could extract rich features from the images with various poses and colors, the vehicles are generally rigid structure with solid colors and appearance is easily affected by various illuminations, viewpoints. Most existing works only focus on learning the discriminative features while neglecting the influence of different cameras. Actually, images captured from different cameras often have obviously different styles. Usually, cameras differ from each other regarding resolution, illumination, background, etc. As Fig. \ref{introduction} shows, for each row, the images with the same identity have different appearances in different camera views. This could lead to serve cross-camera bias and affect the vehicle reID task. Some vehicle reID researches also noticed the challenges, thus preferred to make use of spatial-temporal information and plate license to achieve promising results. However, the spatial-temporal information is usually not annotated in some datasets. Besides that, the high-resolution images of in front or rear viewpoints are required for license plate recognition, which is not impractical in the real-world scenes.

In order to solve these problems, some methods consider learning the global features from multi-view images, such as VAMI  \cite{zhou2018aware} and DHMV \cite{zhou2018vehicle}. VAMI adopts cross-view generative adversarial network to transform the features into a global multi-view feature representation. DHMV aims to learn transformations across different viewpoints for inferring the multi-view representation from one input vehicle image. There are also some methods that exploit the constraint among cross-cameras by proposing the cross-view losses. For instance, MVR loss \cite{lin2019multi} introduced several latent groups to represent multiple views and ranked them by calculating the intra and inter loss. However, these methods only consider the influence of various viewpoints to solve the problem between different cameras and neglect the background and other factors.

The above issues prompt us to focus on the changes of images caused by different cameras. To solve aforementioned problems, this paper proposes a cross-camera adaptation network (CCA) to smooth the bias between different cameras and learn powerful features. In our paper, the single camera is regard as an independent domain. CCA aims at transforming the multi-domains into one common domain that has the similar background, illumination and resolution. Different with existing methods, CCA firstly generates vehicle images by StarGAN, which transfers the same vehicle images from other cameras into one camera and doesn't augment the quantity of original datasets. Besides, it could be observed in the Fig. \ref{introduction}, the images captured from different always have different backgrounds, which may interfere with the training of vehicle reID model. Hence, to eliminate the impact of background, the attention alignment network (AANet) is proposed to locate discriminative features. Specially, the STN with attention module is employed to select a series of regions from vehicle images for training a powerful reID model. The main contributions of our work can be summarized as follows£º
\begin{itemize}
\item A cross-camera adaptation framework is proposed for better smoothing the bias between different cameras, which reduces the influence of illumination, background and resolution for vehicle reID task by transferring images into a common space and learning a powerful discriminative feature.

\item The attention alignment network is proposed to obtain a series of local regions for vehicle reID, which focus on locating the meaningful parts while suppressing background. Moreover, Extensive experiments demonstrate that our proposed method achieves competitive performance on challenging benchmark datasets.

\end{itemize}

The rest of this paper is organized as follows. In Section 2, we review and discuss the related works. Section 3 illustrates the proposed method in detail. Experimental results and comparisons on two vehicle reID datasets are discussed in Section 4, followed by conclusions in Section 5.

\section{Related Works}
In this section, existing vehicle reID works are reviewed. With the prosperity of deep learning, vehicle reID has achieved some progress in recent years. Broadly speaking, these approaches could be categorized into three classes, i.e., representation learning, similarity learning and spatio-temporal correlation learning.

A series of methods attempt to identify vehicles based on the visual appearance. In \cite{zapletal2016vehicle}, 3D bounding boxes of vehicles were detected and then were processed by the color histograms and histograms of oriented gradients for vehicle reID. In \cite{zhao2019structural}, a ROIs-based vehicle reID method was proposed to detect the ROIs' as discriminative identifiers. And then encode the structure information of a vehicle for reID task. DHMVI \cite{zhou2018vehicle} utilized the LSTM bi-directional loop to learn transformations across different viewpoints of vehicles, which could infer all viewpoints' information from the only one input view. RAM \cite{liu2018ram} was proposed for vehicle reID task with several branches including region and attribute branches to extract distinct features from several overlapped local regions. MRM \cite{peng2019learning} introduced a multi-region model to extract features from a series of local regions for learning powerful features for vehicle reID task. EALN \cite{lou2019embedding} was introduced to improve the capability of the reID model by automatically generating hard negative samples in the specified embedding space to train the reID model. VAMI \cite{zhou2018aware} tried to better optimize the reID model by transforming single-view feature into a global multi-view feature representation through generative adversarial network. CV-GAN \cite{zhou2017cross} was conducted to generate the various viewpoints vehicle images by generative adversarial network for training an adaptive reID model.

Apart from the visual appearance, a series of metric losses for deep feature embedding to achieve higher performance. In \cite{liu2016deep}, coupled cluster loss was proposed to push those negative ones far away and pull the positive images closer, which could minimize maximize inter-class distance and intra-class distance to train the vehicle reID model. GST loss \cite{bai2018group} was proposed to deal with intra-class variance in learning representation. Besides that, it introduced the mean-valued triplet loss to alleviate the negative impact of improper triplet sampling during training stage. MGR \cite{guo2019two} was presented to further enhanced the discriminative ability of reID model by enhancing the discrimination that not only between different vehicles but also different vehicle models.

Besides, spatio-temporal information is an important cue for vehicle reID task. Hence, some approaches exploit spatial and temporal information for vehicle images to improve vehicle reID performance. PROVID \cite{liu2017provid} employed visual features, spatial-temporal relations and the information of license plates with a progressive strategy to learn similarity scores between vehicle images. OIFE \cite{wang2017orientation} refined the retrieval results of vehicles by utilizing the log-normal distribution to model the spatio-temporal constrains in camera networks. Siamese-Cnn+Path-LSTM \cite{shen2017learning} model was proposed to incorporate complex spatio-temoral information for regularizing the reID results.

\section{Cross-camera Adaptation Framework}
\begin{figure*}[htbp]
  \includegraphics[width=12cm]{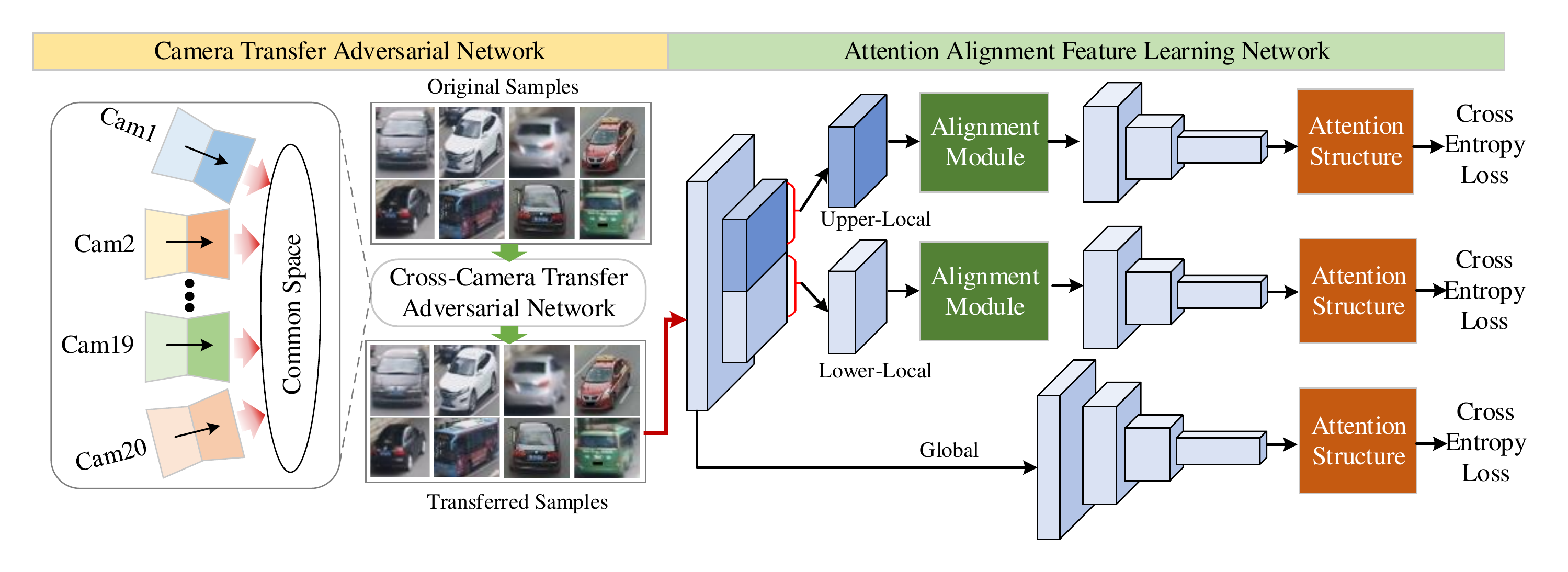}
\caption{Cross-camera Adaptation Framework. The proposed framework is composed of the camera transfer adversarial network and attention alignment network. In the camera transfer adversarial network, the samples in different cameras are transferred into the common space, which means the generated samples have the similar distribution. Subsequently, the transferred samples are employed to train the reID model with the attention alignment network.}
\label{framework}       
\end{figure*}

The overall structure of the proposed framework is depicted in Fig.\ref{framework}. The Cross-camera Adaptation Framework (CCA) is composed of the camera transfer adversarial network and the attention alignment network. Firstly, the samples from different cameras are transferred into one domain by the camera transfer adversarial network. Then the images with similar distribution could be obtained, which are fed into the proposed attention alignment feature learning network for training the reID task. Specially, the attention alignment network is a dual-branches network that focus on different meaningful parts of vehicle images for improving the discriminate ability of reID model.

In this section, we introduce our method from two aspects: 1) an camera transfer adversarial network is introduced in section 3.2, which learns transfer mappings for different cameras; 2) an attention alignment feature learning network AANet is illustrated in section 3.3, which optimizes the reID model utilizing the generated images from the camera transfer adversarial network.

\subsection{Camera Transfer Adversarial Network}
\begin{figure}[htbp]
\centering
  \includegraphics[width=8cm]{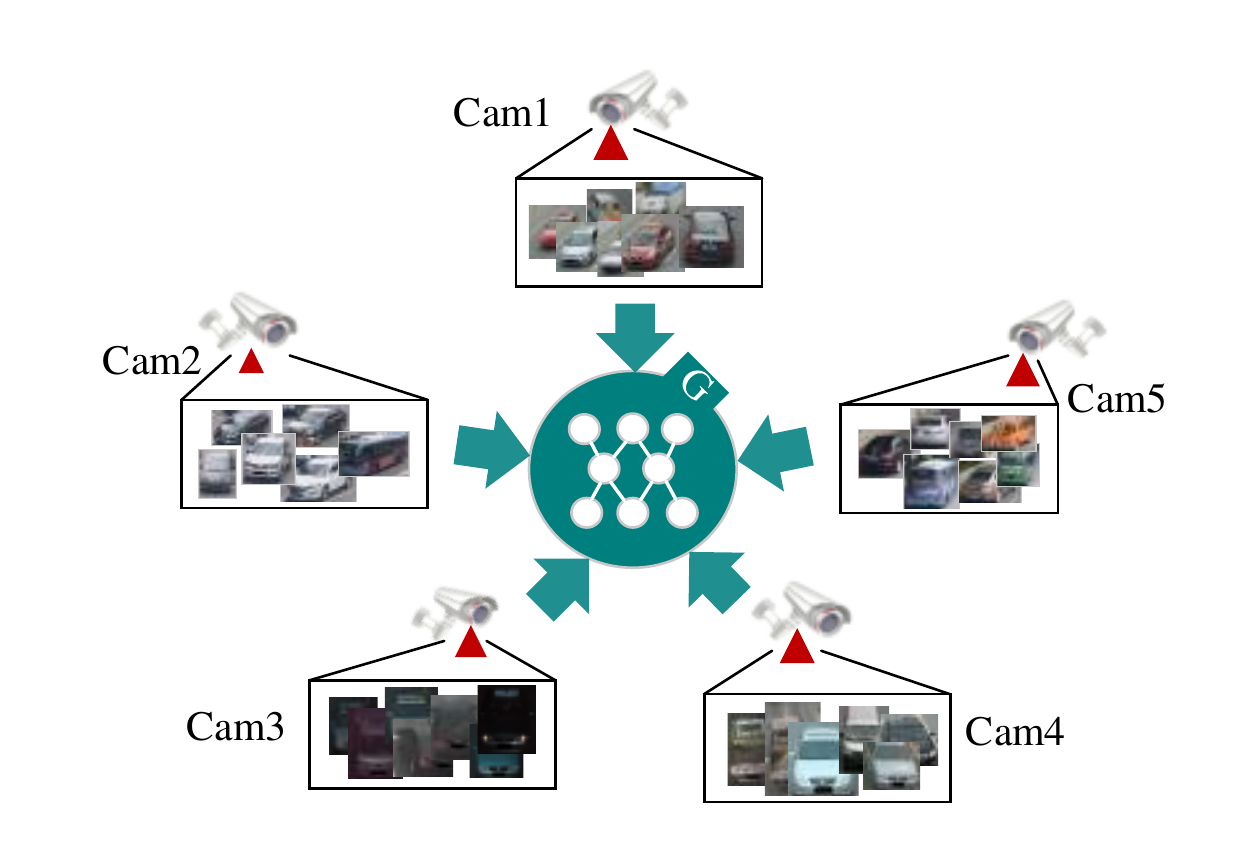}
\caption{The structure of starGAN. In our paper, the vehicle images from different cameras are transferred into common space by starGAN. Thus, these images have the similar distribution, which could be regarded as one common domain.}
\label{starGAN}       
\end{figure}

The same vehicles always have different appearances in different camera views and the bias is shown in Fig.\ref{introduction}. In this paper, to smooth the bias between different cameras, we want to transfer images in different cameras into one camera, which means that all images have the similar distribution. To achieve this, StarGAN \cite{choi2018stargan} is utilized as the camera transfer adversarial network. StarGAN \cite{choi2018stargan} utilizes generator $G$ and discriminator $D$ to implement the conversion between multiple cameras, which learns the mapping relations among multiple cameras using only a single model, as shown in Fig.\ref{starGAN}.

In StarGAN, in order to generate a more realistic fake sample, an adversarial loss function is employed to obtain the high-quality image, which could be written as:
\begin{equation}
{L_{adv}} = {E_x}\left[ {\log {D_{src}}(x)} \right] + {E_{x,c}}\left[ {\log (1 - {D_{src}}(G(x,c)))} \right]
\end{equation}
where $G$ generates an image $G(x,c)$ to fake $D$. $D$ tries to distinguish the real image from the generated image. The target of StarGAN is to translate $x$ to an output images $y$ that is classified as the target domain $c$. For this goal, a domain classifier is added on the $D$, which could be defined as:

\begin{equation}
L_{dom}^r(x,c^*) = {E_{x,{c^*}}}\left[ { - \log {D_{dom}}({c^*}|x)} \right]
\end{equation}

\begin{equation}
L_{dom}^f(x,c) = {E_{x,{c}}}\left[ { - \log {D_{dom}}({c}|x)} \right]
\end{equation}
where ${{D_{dom}}({c^*}|x)}$ is the probability distribution over the camera labels of a given real image $x$, and $c$ represents the source camera labels. To guarantee that generated images could preserve the identity information of original images, StarGAN employs the cycle consistent loss \cite{zhu2017unpaired}, which is defined as:

\begin{equation}
{L_{rec}}(x,c,c^*) = {E_{x,c,{c^*}}}\left[ {{{\left\| {x - G(G(x,c),{c^*})} \right\|}_l}} \right]
\end{equation}

Through StarGAN, one image could be transferred into any other cameras. Hence, there are $N$ times images than original dataset. $N$ is the number of cameras. However, in our paper, we aim to transfer images into one common domain. So we just select images from one camera for training vehicle reID model. As illustrated in Fig.\ref{introduction}, the images with irrelevant background or less discriminative parts of objects of interest may confuse the reID model, which would degenerate the model's performance. To solve this problem, in this paper, AANet is proposed to utilize the style-translated images as training set to guide the reID model to focus on the discriminative parts, to be detailed in Section 3.2.

\subsection{Attention Alignment Network}
\begin{figure*}[htbp]
  \includegraphics[width=12cm]{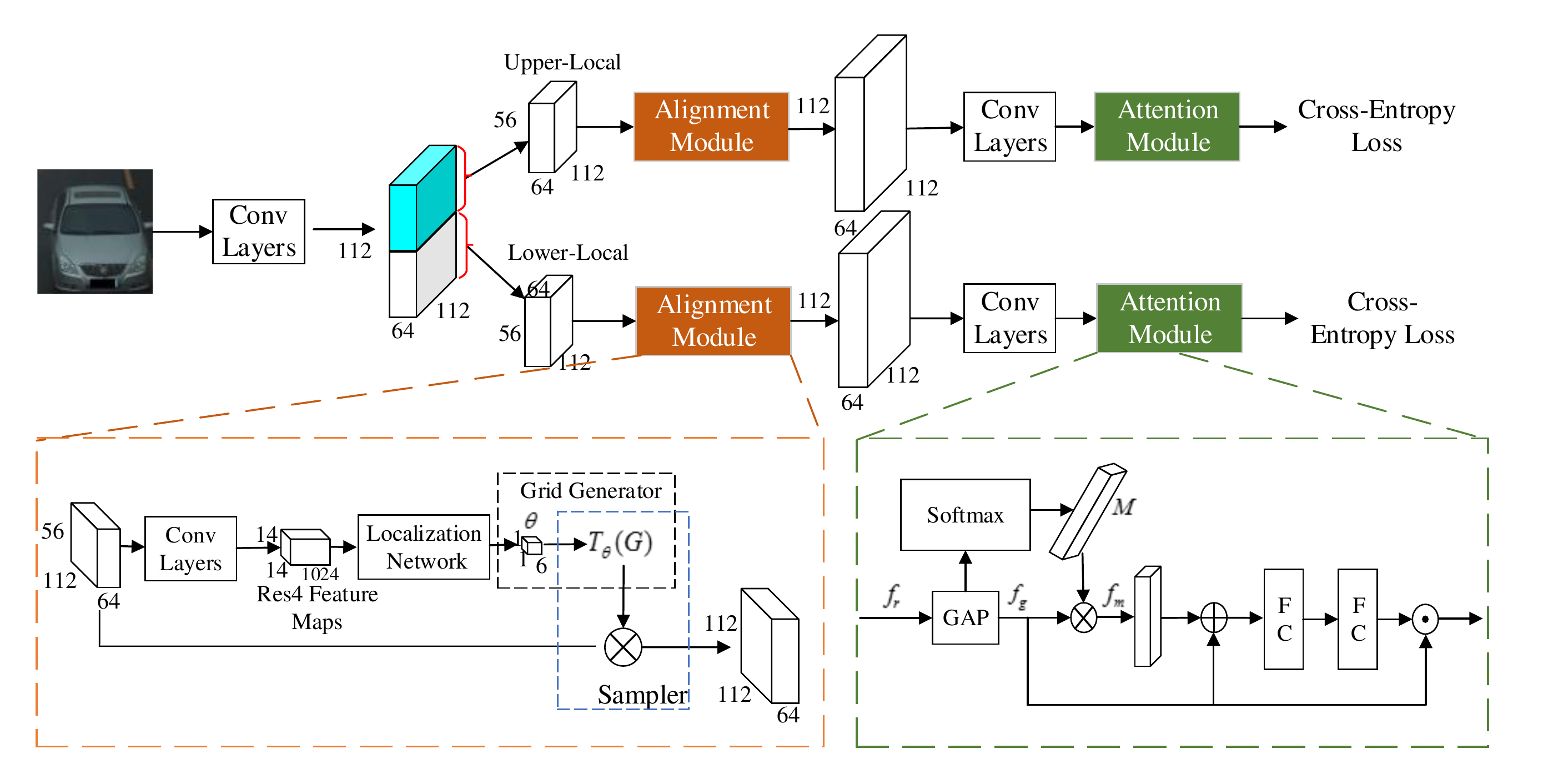}
\caption{The structure of Attention Alignment Network. The network contains three branches. The global branch is utilized to learn the context features. Besides that, for each local branch, we focus on learning features of different regions by the alignment module and attention module. Each of the generated feature from three branches is trained with an individual classifier with cross-entropy loss.}
\label{AANet}       
\end{figure*}

Redundancy of background information is another important factor that obstructs vehicle reID performance. Based on the transferred images, we propose the attention alignment network (AANet) to reduce the discrepancy of attention maps across non-overlapping cameras. The AANet is designed to focus on the meaningful parts of vehicle images and neglect the background when training the reID model, which is illustrated in Fig.\ref{AANet}.

The AANet is designed as a multi-branch structure, which is composed of one global branch and two local region branches. For the global branch, it is utilized to learn the context features with the attention module. Besides that, for the local regions, in order to obtain key information from the local region, we divided the output feature map generated by several convolutional layers with the size of $112 \times 112 \times 64$ into two non-overlapping local regions, which could be named as ``Upper-Local'' and ``Lower-Local'', respectively. Then, the feature maps are fed into two branches to generate different features. So given an input vehicle image, the local region network could generate a series of features for vehicle ReID.

Specially, to address the problems of excessive background and extract the remarkable features, in each branch, an alignment module based on the STN is employed. The alignment module includes three components, a localization network to learn the transformation parameters, a grid generator to calculate the coordinate of the input feature maps by applying the transformation parameters and bilinear sampler to make up the missing pixels. Meanwhile, as shown in Fig.\ref{AANet}, to focus on the meaningful parts of vehicle images and neglect the background when training the feature learning model, an attention module is introduced to generate discriminative features. In the attention module, after a global average pooling layer, we employ the Softmax layer to re-weight the feature maps and generate the mask, which could be computed as:

\begin{equation}
M = Softmax(Conv(GAP(f_{r})))
\end{equation}
where the $Conv$ operator is $1\times 1$ convolution. The $M$ is the weight matrix. After obtaining $M$, the attended feature map could be calculated by $f_m = f_{a}\otimes M$. The operator $\otimes$ is performed in an element-wise product. Then the attended feature map $f_{m}$ is fed into the subsequent structure.

\begin{figure*}[htbp]
  \includegraphics[width=12cm]{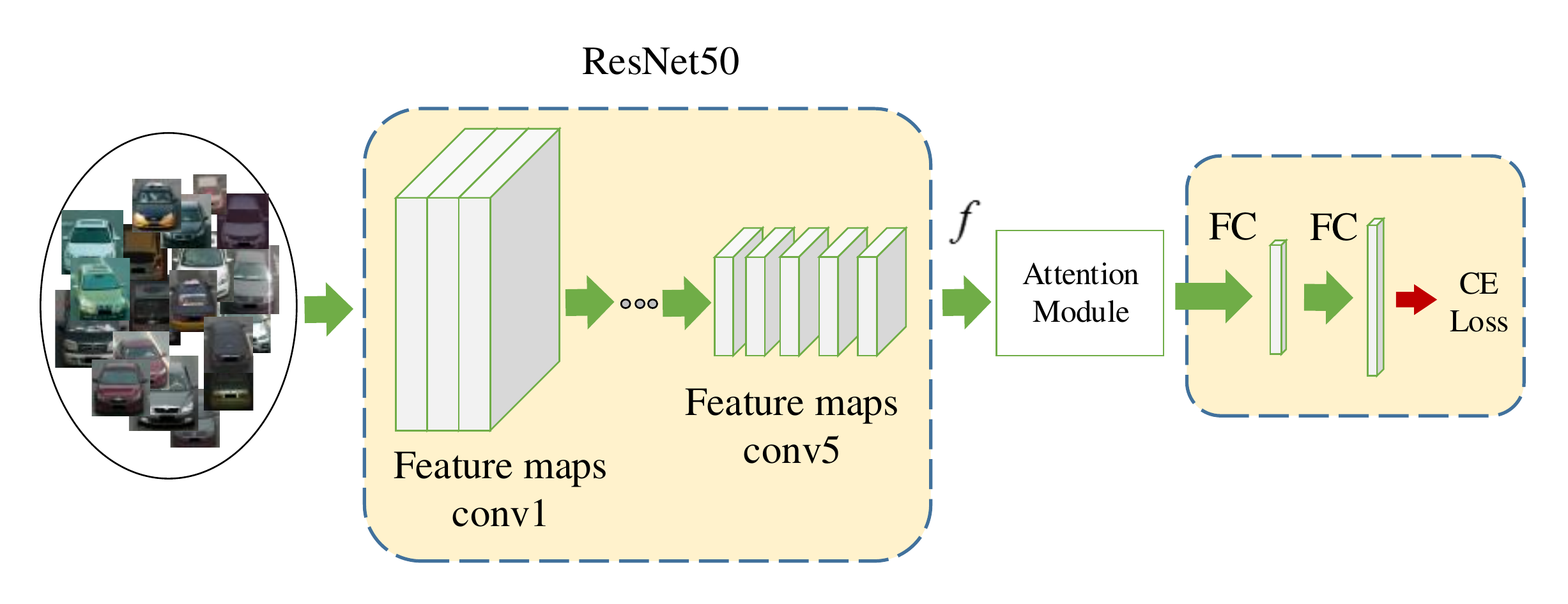}
\caption{One branch of global branch Network. The network adopts the ResNet50 as the base model and employs the attention module for extracting distinct features.}
\label{resnet}
\end{figure*}

The structure of global branch is also a two-branch network that is introduced in \cite{zheng2018discriminatively}. In our paper, for one branch, as shown in Fig.\ref{resnet}, the ResNet50 \cite{he2016deep} is adopted as the base model for vehicle classification, which consists of residual units that preserve the identity and maintain a deeper structure. After convolutional layers from $conv1$ through $conv5$, the feature vector $f$ could be obtained. Similar with the local region branch, the features $f$ is fed into the attention module to obtain distinct features. Then the output feature is utilized to train the identification task with the cross-entropy (CE) loss.

At last, for all branches, the obtained features are named as $f_{g}$, $f_{u}$ and $f_{l}$, respectively. During the training phase of each branch in local region features learning network, Fully Connected (FC) layers are added to identify vehicles only with a part of feature maps as input. This procedure enforces the network to extract discriminative details in each part. At last, the prediction identity classification is given by the FC layer with the CE loss that could be described as:

\begin{equation}
L_{p}(\theta) = \ell_{id}^{g} + \lambda_{1}\ell_{id}^{u} + \lambda_{2}\ell_{id}^{l}
\end{equation}
where $\theta$ denotes the parameters in the deep model. $\ell_{id}^{g}$, $\ell_{id}^{u}$ and $\ell_{id}^{l}$ represent the identification loss in global local region features extraction module, respectively. $\lambda_{1}$ and $\lambda_{2}$ are the weights for corresponding loss.

The CE loss is calculated based on softmax which is formulated as follows:
\begin{equation}
p = -\sum_{i=1}^{m}log\frac{e^{{W_{y_{i}}^{T}x_{i}}+b_{y_{i}}}}{\sum_{j=1}^{n}e^{W_{j}^{T}+b_{j}}}
\end{equation}
where $x_{i}$ is the $i$-th deep feature that belongs to the $y_{i}$-th class. For different datasets, $m$ represents the size of mini-batch and $n$ is the number of class of training set. $d$ is the dimension of the output feature. $b$ represents the bias term and $W_{j}$ denotes the $j$-th column of the weights \cite{ref_article30}.

Specially, during the test phase, the final features from AANet could be described as follows:

\begin{equation}
f = [f_{g}\times \alpha,f_{u}\times(1-\alpha),f_{l}\times(1-\alpha)]
\end{equation}
where $\alpha$ is the weight for features. The size of features from different branches is $1\times1\times4096$ in our paper.

\section{Experiments}

In this section, we evaluate our proposed method for vehicle ReID using the Cumulative Match Characteristic (CMC) curve and mean Average Precision (mAP) \cite{lin2019improving,wu2019few} widely adopted in vehicle ReID. Besides comparing with state-of-the-art vehicle ReID methods, a series of detailed studies are conducted to explore the effectiveness of proposed method. All the experiments are conducted on two vehicle ReID datasets: VeRi-776 \cite{liu2017provid} and VehicleID \cite{liu2016deep}.

\subsection{Datasets and Evaluation Metrics}

\begin{itemize}
\item VeRi-776  \cite{liu2017provid}. The dataset is a large-scale urban surveillance vehicle dataset for reID, which contains over 50,000 images of 776 vehicles across 20 cameras. Each vehicle is from 2-18 cameras with various viewpoints, illuminations and occlusions. In this dataset, 37,781 images of 576 vehicles are split as a train set and 11,579 images of 200 vehicles are employed as a test set. A subset of 1,678 images in the test set generates the query set.

\item VehicleID \cite{liu2016deep}. It is a widely-used vehicle reID dataset, which contains 26267 vehicles and 221763 images in total. The training set contains 110,178 images of 13,134 vehicles. For the testing data, three subsets which contain 800, 1600, and 2400 vehicles are extracted for vehicle search in different scales. During testing phase, one image is randomly selected from one class to obtain a gallery set with 800 images, then the remaining images are all utilized as probe images. Two other test sets are processed in the same way.

\item Evaluation Metrics. To measure the performance for vehicle reID task, the CMC and mAP \cite{lin2019improving} are utilized as evaluation criterions. For each query, its average precision (AP) is computed from its precision-recall curve. And mAP is the mean value of average precisions across all queries.
\end{itemize}

\subsection{Implementation Details}
For the translation module, the model is trained in the pytorch \cite{paszke2017automatic}. We utilize the Adam optimizer \cite{li2014deepreid} with $\beta_1=0.5$ and $\beta_2=0.999$. The initial learning rate is 0.0001 for the first 100 epochs and linearly decays to the learning to 0 over the next 100 epochs. The batch size is 16. For the feature learning network, we implement the proposed vehicle reID model in the Matconvnet \cite{van2014accelerating} framework. SGD  \cite{bottou2010large} is employed to update the parameters of the network with with a momentum of $\mu=0.0005$ during the training procedure on both VehicleID and VeRi-776. The batch size is set to 16. Besides that, the learning rate of the first 40 epochs is set to 0.1 while the last 25 is 0.01.

\subsection{Comparison with the state-of-the-art methods}

\subsubsection{Comparison on VeRi-776} The results of the proposed method is compared with state-of-the-art methods on VeRi-776 dataset in Tables \ref{tab1} \ref{tab2}, which includes: (1) LOMO \cite{liao2015person}; (2) DGD \cite{xiao2016learning}; (3) GoogLeNet \cite{yang2015large} (4) FACT+Plate-SNN+STR \cite{liu2016deep2}; (5) NuFACT+Plate-REC \cite{liu2017provid}; (6) PROVID \cite{liu2017provid}; (7) Siamese-Visual \cite{shen2017learning}; (8) Siamese-Visual+STR \cite{shen2017learning}; (9) Siamese-CNN+Path-LSTM \cite{shen2017learning}; (10) OIFE+ST \cite{wang2017orientation}; (11) VAMI \cite{zhou2018aware}; (12) VAMI+ST \cite{zhou2018aware}. From the Tables \ref{tab1} \ref{tab2}, it should be noted that the proposed method achieves the best performance among the compared with methods with rank-1 = 91.71\%, mAP = 68.05\% on VeRi-776, which acquires the highest mAP and rank-1 among all methods under comparisons. More details are analyzed as follows.

Firstly, the proposed AGNet obtains much better performance than those hand-crafted feature representation methods, such as LOMO \cite{liao2015person} and DGD \cite{xiao2016learning}, which achieves 58.41 and 50.13 points in mAP improvements, respectively. This verifies that the features obtained from deep model are more robust than the hand-crafted feature that are severely affected by the complicated environment.

Second, compared with those methods that learn multi-view features, the proposed also show satisfactory performance. For instance, compared with VAMI, our method has a gain of 17.92 in terms of mAP and 14.68 in terms of rank-1 accuracy. This is because that our method eliminates background interference information. It strongly proves that the bias between camera has a serve influence on the vehicle reID task.

Thirdly, although our proposed method only utilizes visual information, it also has significant improvements when compared with methods with spatio-temproal information. such as FACT+Plate-SNN+STR \cite{liu2016deep2}, PROVID \cite{liu2017provid}, Siamese-Visual+STR \cite{shen2017learning}, Siamese-CNN+Path-LSTM \cite{shen2017learning}, OIFE+ST \cite{wang2017orientation} and VAMI+ST \cite{zhou2018aware}, the proposed method has higher mAP, rank-1 and rank-5 than them, which demonstrates that our AGNet could extract more discriminative features without other information besides the vehicle images.

\begin{table}[htbp]
\renewcommand{\arraystretch}{1.3}
\centering
\caption{Experimental results on VeRi-776. The mAP (\%) and cumulative matching scores (\%) at rank 1, 5 are listed.}\label{tab1}
\begin{tabular}{p{4.3cm}|p{1.0cm}|p{1.0cm}|p{1.0cm}}
\hline
 Method & mAP & Rank1 & Rank5\\
\hline
\hline
LOMO \cite{liao2015person} &  9.64 & 25.33 & 46.48\\
DGD \cite{xiao2016learning} &  17.92 & 50.70 & 67.52\\
GoogLeNet \cite{yang2015large} &  17.81 & 52.12 & 66.79\\
FACT+Plate-SNN+STR \cite{liu2016deep2} &  27.77 & 61.64 & 78.78\\
NuFACT+Plate-REC \cite{liu2017provid} &  48.55 & 76.88 & 91.42\\
PROVID \cite{liu2017provid} &  53.42 & 81.56 & 95.11\\
Siamese-Visual \cite{shen2017learning} & 29.48 & 41.12 & 60.31\\
Siamese-Visual+STR \cite{shen2017learning} & 40.26 & 54.23 & 74.97\\
Siamese-CNN+Path-LSTM \cite{shen2017learning} & 58.27 & 83.49 & 90.04\\
OIFE+ST \cite{wang2017orientation} & 51.42 & 68.30 & 89.70\\
VAMI \cite{zhou2018aware} & 50.13 & 77.03 & 90.82\\
VAMI+ST \cite{zhou2018aware} & 61.32 & 85.92 & 91.84\\
\hline
\hline
CCA &  68.05 & 91.71 & 96.90\\
\hline
\end{tabular}
\end{table}

\begin{table*}[htbp]
\renewcommand{\arraystretch}{1.3}
\centering
\caption{Experimental results on VehicleID. The mAP (\%) and cumulative matching scores (\%) at Rank 1, 5 are listed.}\label{tab2}
\begin{tabular}{p{2.1cm}|p{0.7cm}|p{0.8cm}|p{0.8cm}|p{0.7cm}|p{0.8cm}|p{0.8cm}|p{0.7cm}|p{0.8cm}|p{0.8cm}}
\hline
 \multirow{2}*{  Method} & \multicolumn{3}{c|}{ Test size = 800} & \multicolumn{3}{c|}{ Test size = 1600}& \multicolumn{3}{c}{ Test size = 2400} \\ \cline{2-10} &  mAP & Rank1 & Rank5 & mAP & Rank1	& Rank5 & mAP & Rank1 & Rank5 \\
\hline
\hline
BOW-SIFT \cite{liu2016deep2}   & -  & 2.81  & 4.23 &- & 3.11 & 5.22 & -	& 2.11	& 3.76\\
LOMO \cite{liao2015person}  & -  & 19.76  & 32.14 &- & 18.95 & 29.46 & -	& 15.26	& 25.63\\
DGD \cite{xiao2016learning}  & -  & 44.80  & 66.28 &- & 40.25 & 65.31 & -	& 37.33	& 57.82\\
VGG+T \cite{liu2016deep}    & -  & 40.4  & 61.7 &- & 35.4 & 54.6 & -	& 31.9	& 50.3\\
VGG+CCL \cite{liu2016deep}  & -	&43.6	&64.2	&-	&42.8	&66.8	&-	&32.9	&53.3\\
Mixed DC \cite{liu2016deep}   & -	&49.0	&73.5	&-	&42.8	&66.8	&-	&38.2&	61.6\\
FACT  \cite{liu2017provid}  & -	&49.53	&67.96	&-	&44.63	&64.19	&-	&39.91&	60.49\\
NuFACT \cite{liu2017provid}  & -	&48.90	&69.51	&-	&43.64	&65.34	&-	&38.63&	60.72\\
OIFE \cite{wang2017orientation}   &-	&-&	-	&-	&-	&-	&-	&67.0	&82.9\\
VAMI \cite{zhou2018aware} &-	&63.12&	83.25	&-	&52.87	&75.12	&-	&47.34	&70.29\\
TAMR \cite{guo2019two}    &67.64	&66.02 &79.71	&63.69	&62.90	&76.80	&60.97	&59.69	&73.87\\
\hline
\hline
CCA    &78.89 &75.51	&91.14 &76.53 &73.60 &86.46	&73.11 &70.08 &83.20\\
\hline
\end{tabular}
\end{table*}

\subsubsection{Comparison on VehicleID} There are 9 methods are compared with our proposed method, which are (1) LOMO \cite{liao2015person}; (2) DGD \cite{xiao2016learning}; (3) VGG+T \cite{liu2016deep}; (4) VGG+CCL \cite{liu2016deep}; (5) Mixed DC \cite{liu2016deep}; (6) FACT \cite{liu2017provid}; (6) NuFACT \cite{liu2017provid}; (7) OIFE \cite{wang2017orientation}; (8) VAMI \cite{zhou2018aware}; (9) TAMR \cite{guo2019two}. Table .. illustrates the rank-1, rank-5 and mAP of our method and other comparison methods on VehicleID. Firstly, it can be observed that deep learning based methods obviously outperform traditional methods. And compared with traditional methods LOMO \cite{liao2015person} and DGD \cite{xiao2016learning}, the proposed method has 55.75\% and 30.71\% gains in rank-1 on the test set 800, respectively. The similar improvements also occur on other test sets. Secondly, Different VeRi-776, there is no spatio-temporal labels in VehicleID. Hence, there are no methods that consider the spatio-temporal information. All compared methods utilize the appearance information only from vehicle images. The proposed method outperforms all deep learning based methods under comparison on the test sets with different sizes on VehicleID, which obtains 75.51\%, 73.60\%, 70.08\% in rank-1, respectively.  And this also shows that our proposed method could generate more distinct features for different vehicle reID datasets.


\subsection{Evaluation of proposed method}

To validate the necessity of the proposed method, some ablation experiments are conducted. The comparison results on VeRi-776 and VehicleID are presented in Table \ref{tab3} and Table \ref{tab4}. ``Original'' means the the training set is original samples while ``Transfer'' is the generated samples. ``Rigid'' represents the training network doesn't employ the STN module and attention module, which is divided into two parts from resnet50 directly. ``Part-n'' is the descriptor of $i$-th branch. ``global'' means the descriptor is only composed of the features from global branch.

\begin{table}[htbp]
\renewcommand{\arraystretch}{1.3}
\centering
\caption{Performance of features fusion on VeRi-776. The mAP (\%) and cumulative matching scores (\%) at Rank 1, 5 are listed.}\label{tab3}
\begin{tabular}{p{3.5cm}|p{1.0cm}|p{1.0cm}|p{1.0cm}}
\hline
 Descriptor & mAP & Rank1 & Rank5\\
\hline
\hline
global &  56.71 & 86.55 & 92.14\\
\hline
original-Rigid-Part1 &  48.15 & 81.58 & 90.04\\
original-Rigid-Part2 &  47.51 & 81.10 & 90.88\\
original-Rigid-All &  61.19 & 87.24 & 93.32\\
\hline
original-AANet-Part1 &  50.77 & 83.07 & 92.01\\
original-AANet-Part2 &  50.41 & 82.24 & 92.19\\
original-AANet-All &  65.45 & 89.92 & 94.39\\
\hline
transfer-AANet-Part1 & 54.04 & 86.94 & 93.74\\
transfer-AANet-Part2 &  53.78 & 86.59 & 93.98\\
transfer-AANet-All &  68.05 & 91.71 & 96.90\\
\hline
\end{tabular}
\end{table}

\begin{table*}[htbp]
\renewcommand{\arraystretch}{1.3}
\centering
\caption{Performance of features fusion on VehicleID. The mAP (\%) and cumulative matching scores (\%) at Rank 1, 5 are listed.}\label{tab4}
\begin{tabular}{p{3.5cm}|p{0.8cm}|p{0.8cm}|p{0.8cm}|p{0.8cm}|p{0.8cm}|p{0.8cm}}
\hline
\multirow{2}*{ Descriptor} & \multicolumn{3}{c|}{ Test size = 800} & \multicolumn{3}{c}{Test size = 1600}\\
\cline{2-7} & mAP & Rank1 & Rank5 & mAP & Rank1 & Rank5 \\
\hline
 global &69.78 &66.51 &79.25 &67.71 &64.79 &78.86 \\
\hline
 original-Rigid-Part1    &67.82 &65.69 &74.28 &65.37 &63.38 &71.22\\
 original-Rigid-Part2     &67.43 &65.25 &73.98 &64.64 &62.60 &72.57\\
 original-Rigid-All         &74.93 &71.45 &88.32 &72.36 &69.34 &82.41\\
\hline
 original-AANet-Part1   &69.22 &66.69	&77.00 &67.44 &64.81 &75.37\\
 original-AANet-Part2     &70.64 &67.97	&79.15 &68.57 &64.11 &74.19\\
 original-AANet-All        &77.27 &73.79	&89.98 &74.47 &71.07 &84.90\\
\hline
 transfer-AANet-Part1   &71.78 &69.47	&79.43 &70.17 &68.06 &76.60\\
 transfer-AANet-Part2     &71.24 &68.87	&78.93 &70.10 &67.91 &76.67\\
 transfer-AANet-All    &78.89 &75.51	&91.14 &76.53 &73.60 &86.46 \\

\hline
\multirow{2}*{ Descriptor} & \multicolumn{3}{c|}{Test size = 2400} & \multicolumn{3}{c}{ Test size = 3200} \\
\cline{2-7} & mAP & Rank1 & Rank5 & mAP & Rank1 & Rank5 \\
\hline
 global &64.43 &60.68 &74.37 &62.88 &59.79 &71.53 \\
\hline
 original-Rigid-Part1    &63.42 &61.50 &68.85 &61.96 &60.21 &66.78\\
 original-Rigid-Part2      &63.48 &61.66 &68.54&61.69 &59.94 &66.61\\
 original-Rigid-All     &67.58 &63.60 &81.01 &67.91 &65.21 &75.71\\
\hline
 original-AANet-Part1   &65.30 &62.75	&73.07 &65.12 &63.13 &77.35\\
 original-AANet-Part2     &64.77 &62.45	&71.62 &63.17 &60.96 &69.50\\
 original-AANet-All    &71.19 &68.01	&81.54 &69.11 &66.18 &78.17\\
\hline
 transfer-AANet-Part1   &68.04 &66.06	&73.85 &66.12 &64.23 &71.40\\
 transfer-AANet-Part2     &67.65 &65.62	&73.64 &66.16 &64.28 &71.36\\
 transfer-AANet-All     &73.11 &70.08 &83.20 &70.75 &67.98 &79.35 \\
\hline

\end{tabular}
\end{table*}

Firstly, the difference of ``Original-AANet-All'' and ``Transfer-AANet-All'' is only the source images of training sets. Hence, compared with ``Original-AANet-All'', the ``Transfer-AANet-All'' has gains of 2.6\%, 1.65\% in mAP and rank-1 on VeRi-776, which demonstrates that through the cross-camera transfer network, the bias of different cameras has dropped. Besides that, because our descriptor is learned by multiple branches in the proposed network, we design an ablation experiment analyzing the effectiveness of global, part and fusion feature. ``Transfer-AANet-All'' is our proposed method that combines all features for reID task. ``Transfer-AANet-Part1'' and ``Transfer-AANet-Part2'' denote the features are extracted by the upper branch and lower branch, respectively. As reported in Table \ref{tab3} and Table \ref{tab4}, it is worth noting that, for each group, the match rates of all independent features are lower than the combination features, such as the ``global'', ``Transfer-AANet-Part1''`and , ``Transfer-AANet-Part2''. However, the match rate further increases slightly when adding the part features and global features. For instance, on VeRi-776, compared with ``global'', ``Transfer-AANet-All'' improves 11.34\% in mAP. It shows that combining with global and part feature can provide more useful information.

To verify the effectiveness of localization model, we remove the STN module and attention module in the AANet and divide the vehicle image into two regions directly as rigid parts. On VeRi-776, compared with ``original-Rigid-All'', ``original-AANet-all'' has gains of 2.34\%, 2.68\% in mAP and rank-1, respectively. For VehicleID, we also observe improvements of 2.34\%, 2.11\%, 3.61\%, 1.2\% in mAP on test set with the size of 800, 1600, 2400 and 3200. All of these show that the proposed AANet could learn more discriminative features for vehicle reID.


\subsection{Visualization of Results}

\begin{figure*}[htbp]
\includegraphics[width=12.5cm]{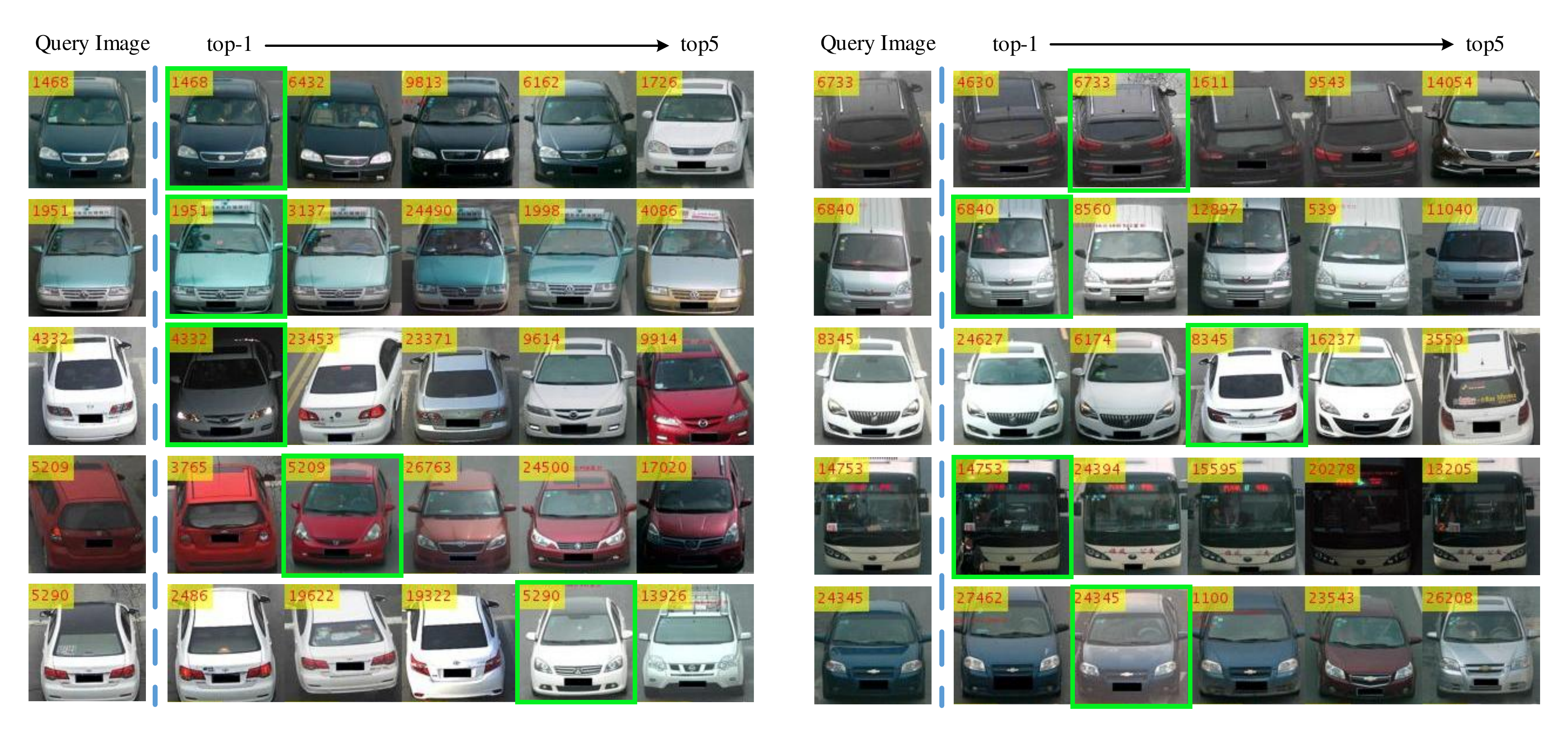}
\caption{The retrieval results on the VehicleID. The left column shows query images while the images of right-hand side are retrieval results obtained by proposed method} \label{fig10}
\end{figure*}

Furthermore, to illustrate the validate of the proposed CCA, some experiment results on VehicleID are visualized. Examples are shown in Fig.\ref{fig10}. In Fig.\ref{fig10}, There are two group results on VehicleID. For each group, the left column shows query images, while images on the right-hand side are the top-5 results obtained by the proposed CCA. Vehicle images with green border are right results while other images are wrong results. For all results, the number on the left-top means Vehicle ID. The same Vehicle ID represents the same vehicle. The Camera ID is the camera number that images are captured. From Fig.\ref{fig10}, it is significant that our proposed CCA has high accuracy and good robustness to different viewpoints and illumination.

\section{Conclusion}

In this paper, we propose cross-camera adaptation framework for better smoothing the bias between different cameras, which reduces the influence of illumination, background and resolution for vehicle reID task by transferring images into a common space and learning a powerful discriminative feature. Besides that, AANet is designed to obtain a series of local regions for vehicle reID, which focuses on locating the meaningful parts while suppressing background. In this paper, it could be observed that appearances of various viewpoints are totally different, which has a big impact on training the reID model. Hence, in our feature studies, we aim to focus on the extension of dataset that utilizes the generative adversarial network to generate the various viewpoints of vehicle images to improve the performance of reID model.

\section{Acknowledgements}
This work was supported in part by the National Natural Science Foundation of China Grant 61370142 and Grant 61272368, by the Fundamental Research Funds for the Central Universities Grant 3132016352, by the Fundamental Research of Ministry of Transport of P. R. China Grant 2015329225300, by the Dalian Science and Technology Innovation Fund 2018J12GX037 and Dalian Leading talent Grant, by the Foundation of Liaoning Key Research and Development Program, China Postdoctoral Science Foundation 3620080307.


\bibliographystyle{spmpsci}      

\bibliography{mybibfile}

\end{document}